# Boosting multi-demographic federated learning for chest radiograph analysis using general-purpose self-supervised representations

Mahshad Lotfinia (1), Arash Tayebiarasteh (2), Samaneh Samiei (3), Mehdi Joodaki (4), Soroosh Tayebi Arasteh (1,5,6,7)

(1) Department of Diagnostic and Interventional Radiology, University Hospital RWTH Aachen, Aachen, Germany.
(2) Department of Computer Engineering, Hamedan University of Technology, Hamedan, Iran.
(3) Quantitative Cell Dynamics and Translational Systems Biology, University Hospital RWTH Aachen, Aachen, Germany.
(4) Institute for Computational Genomics, Joint Research Center for Computational Biomedicine, University Hospital RWTH Aachen, Aachen, Germany.
(5) Pattern Recognition Lab, Friedrich-Alexander-Universität Erlangen-Nürnberg, Erlangen, Germany.
(6) Department of Urology, Stanford University, Stanford, CA, USA.
(7) Department of Radiology, Stanford University, Stanford, CA, USA.

**Abstract**
Reliable artificial intelligence (AI) models for medical image analysis often depend on large and diverse labeled datasets. Federated learning (FL) offers a decentralized and privacy-preserving approach to training but struggles in highly non-independent and identically distributed (non-IID) settings, where institutions with more representative data may experience degraded performance. Moreover, existing large-scale FL studies have been limited to adult datasets, neglecting the unique challenges posed by pediatric data, which introduces additional non-IID variability. To address these limitations, we analyzed n=398,523 adult chest radiographs from diverse institutions across multiple countries and n=9,125 pediatric images, leveraging transfer learning from general-purpose self-supervised image representations to classify pneumonia and cases with no abnormality. Using state-of-the-art vision transformers, we found that FL improved performance only for smaller adult datasets ($P<0.001$) but degraded performance for larger datasets ($P\leq0.063$) and pediatric cases ($P=0.242$). However, equipping FL with self-supervised weights significantly enhanced outcomes across pediatric cases ($P=0.031$) and most adult datasets ($P\leq0.007$), except the largest dataset ($P=0.052$). These findings underscore the potential of easily deployable general-purpose self-supervised image representations to address non-IID challenges in clinical FL applications and highlight their promise for enhancing patient outcomes and advancing pediatric healthcare, where data scarcity and variability remain persistent obstacles.

# 1. Introduction

Artificial intelligence (AI) has emerged as a transformative tool in medical image analysis[1–3], with the potential to automate and enhance diagnostic accuracy across diverse clinical settings. However, developing reliable AI models requires access to large and diverse labeled datasets, which is particularly challenging in healthcare due to concerns over data privacy, variability across institutions, and limited availability of labeled pediatric datasets[4]. To address these challenges, privacy-preserving and collaborative methods have been developed, allowing institutions to work together without compromising patient confidentiality. Federated learning (FL)[5–9] is a widely adopted decentralized framework that enables multiple institutions to train AI models locally on their data while only sharing model updates for aggregation into a global model[10,11]. This approach not only preserves data privacy but also allows leveraging diverse datasets from different institutions[12].

Despite its promise[13], FL faces major challenges in medical imaging[14], particularly in highly non-independent and identically distributed (non-IID)[6,15–18] settings. These challenges arise from variations in labeling systems[19–21], imaging equipment, patient demographics (such as age, gender, and race), the expertise of labeling clinicians, and dataset sizes[22]. Such disparities often lead to imbalanced contributions from participating institutions, causing the global model to underperform, particularly for datasets with unique distributions[23]. Research has shown that in non-IID settings, institutions with larger sizes ofdatasets may derive limited benefits from FL and, in some cases, even experience performance degradation[22].

The issue of non-IID variability is especially pronounced in pediatric medical imaging[24]. Pediatric chest X-ray analysis differs substantially from adult cases due to anatomical and physiological differences, variations in disease prevalence and progression, and the smaller size of pediatric datasets[25–27]. The scarcity of pediatric data reflects the lower incidence of certain conditions in children and logistical challenges in acquiring labeled data, which further amplify non-IID variability. Consequently, most FL research in medical imaging has focused on adult datasets, leaving pediatric applications underexplored[28,29].

To mitigate the scarcity of pediatric data, a common approach is to combine pediatric datasets with adult data during training, leveraging the abundance of adult cases. However, this strategy often leads to models that fail to generalize effectively to pediatric cases. The pronounced anatomical, physiological, and clinical differences between adult and pediatric populations introduce additional non-IID variability, exacerbating the existing challenges in FL. Chest radiography, as the most commonly performed imaging examination globally[30,31], has led to the availability of numerous public adult chest X-ray datasets[25,32–36] and likely even more labeled private data[37,38]. While FL provides a promising framework to utilize these datasets in a privacy-preserving manner, the highly non-IID nature of real-world settings often limits its effectiveness.

These limitations underscore the urgent need for innovative methods to address the non-IID challenges in FL, enabling robust and equitable diagnostic performance across diverse populations in chest X-ray analysis[22]. While some prior studies have attempted to tackle this issue, they often lacked generalizability due to limitations such as small dataset sizes, low diversity in data, or narrow focus on specific aspects of non-IID variability[39–42]—such as either combining pediatric and adult datasets[43] or exclusively focusing on adult datasets[22,44].

In this study, we address these gaps by leveraging more than 400,000 frontal chest X-rays collected from highly non-IID settings, encompassing diverse populations from across the globe, including both adult and pediatric datasets. Our objective is to mitigate the non-IID challenges in FL for chest X-ray analysis, focusing on the detection of pneumonia and radiographs without abnormalities. Using vision transformers[45]—one of the state-of-the-art architectures for chest X-ray classification—we first systematically analyze the effects of non-IID variability, demonstrating that the pediatric domain itself represents a significant non-IID factor when combined with adult datasets.

### (a) Non-IID data
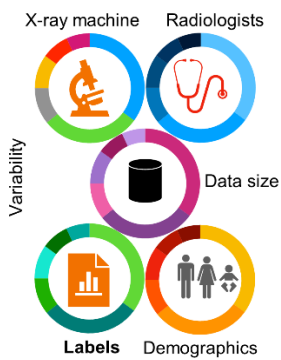

### (b) Conventional FL
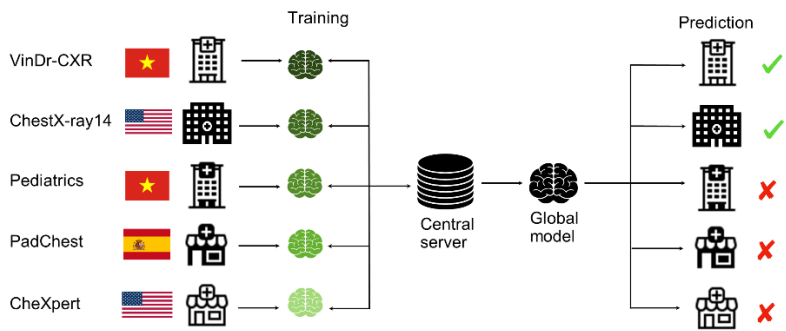

### (c) SSL training
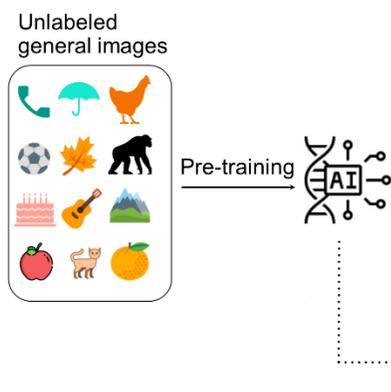

### (d) SSL-based FL
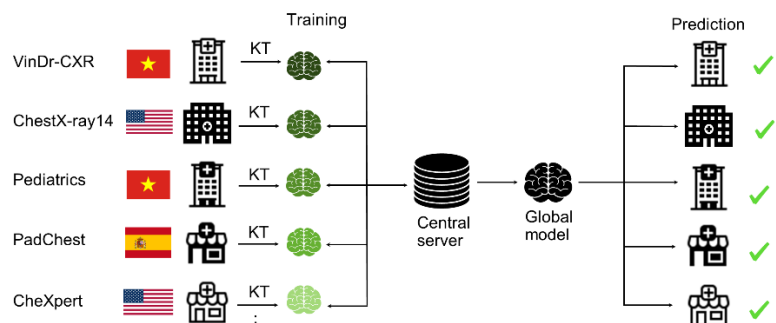

**Figure 1**: **Methodology overview.** (**a**) In real-world scenarios, medical data across institutions are highly non-independent and identically distributed (non-IID) due to factors such as variations in demographics, imaging equipment, labeling protocols, and clinical practices. (**b**) The conventional federated learning (FL) process often struggles in highly non-IID settings, where institutions with representative datasets derive limited benefit from collaboration. (**c**) The general self-supervised learning (SSL) paradigm leverages unlabeled non-medical images to train foundation AI models, utilizing freely available data and bypassing the need for costly manual labeling. (**d**) The SSL-based FL process equips each local institution with SSL weights during the FL process, substantially enhancing the global model's performance and mitigating non-IID effects.

To overcome these challenges, we propose utilizing general-purpose image representations derived from self-supervised learning (SSL) based on the DINOv2 method[46]. We hypothesize that (i) SSL-based FL will improve the generalization of pediatric data within FL frameworks dominated by adult datasets, and (ii) equipping local sites with general-purpose SSL weights will substantially mitigate the broader non-IID effects of conventional FL in chest X-ray analysis (see **Figure 1**). This approach aims to enhance FL's scalability and effectiveness across diverse and highly heterogeneous medical datasets.

**Table 1**: **Overview of the characteristics of the utilized datasets in this study**. This table outlines the included datasets— the Pedi-CXR dataset (Pediatrics), VinDr-CXR, ChestX-ray14, PadChest, and CheXpert— along with their key features. The study utilized only frontal chest radiographs. Note that multiple radiographs from the same patient may be included.

|  | Pediatrics | VinDr-CXR | ChestX-ray14 | PadChest | CheXpert |
|---|---|---|---|---|---|
| Number of Images [n] Total Training set Test set | 9,125 7,728 1,397 | 18,000 15,000 3,000 | 112,120 86,524 25,596 | 110,525 88,480 22,045 | 157,878 128,356 29,320 |
| Patient Age [years] Median | 2 | 57 | 49 | 62 | 61 |
| Female Ratio [%] Training set Test set | 42% 41% | 48% 44% | 42% 42% | 50% 48% | 41% 39% |
| Male Ratio [%] Training set Test set | 58% 59% | 52% 56% | 58% 58% | 50% 52% | 59% 61% |
| Labels [%] Images with pneumonia Images without abnormality (no finding label) | 12% 66% | 4% 70% | 1% 54% | 5% 33% | 2% 11% |
| Image Views [%] Anteroposterior Posteroanterior | 0% 100% | 0% 100% | 40% 60% | 17% 83% | 84% 16% |
| Data Collection Country Period (year) | Vietnam 2020 to 2021 | Vietnam 2018 to 2020 | USA 1992 to 2015 | Spain 2009 to 2017 | USA 2002 to 2017 |

# 2. Results

A total of 407,648 frontal chest radiographs from global institutions were analyzed, encompassing patients with ages ranging from less than 6 months to 105 years. The median ages for the adult datasets were 57 years (VinDr-CXR[32], n=18,000), 49 years (ChestX-ray14[36], n=112,120), 62 years (PadChest[34], n=110,525), and 61 years (CheXpert[33], n=157,878). In contrast, the pediatric dataset[25] (n=9,125) had a median age of 2 years. Characteristics of all datasets are summarized in **Table 1**.

## 2.1. Federated learning shows greater benefits for smaller adult chest X-ray datasets in non-IID settings

We first modeled a realistic and commonly observed scenario in which each institution independently trains a diagnostic model locally for the classification of pneumonia and radiographs with no abnormality, using only its own training data without collaboration. The training data sizes were as follows: n=7,728 (Pediatrics), n=15,000 (VinDr-CXR), n=86,524 (ChestX-ray14), n=88,480 (PadChest), and n=128,356 (CheXpert). These locally trained models were compared with a conventional FL scenario, where each dataset served as an independent local site.

In the conventional FL process, the average area under the receiver operating characteristic curve (AUROC) was 93.67% ± 0.62 (95% CI: 92.74, 94.56) for VinDr-CXR and 71.52% ± 0.79 (95% CI: 70.42, 72.62) for ChestX-ray14. In comparison, the locally trained models achieved an average AUROC of 91.15% ± 0.90 (95% CI: 89.92, 92.32) for VinDr-CXR and 69.20% ± 1.24 (95% CI: 68.00, 70.45) for ChestX-ray14 (see **Table 2**).
As illustrated in **Figure 2**, the conventional FL model significantly outperformed the local models in terms of average AUROC [$P < 0.001$ for both]. These results demonstrate the potential of FL to enhance performance even in highly non-IID settings, particularly for underrepresented adult datasets.

## 2.2. Federated learning performance declines with increasing non-IID effects in chest X-ray datasets

As shown in **Figure 3**, for the Pediatrics dataset, the conventional FL process resulted in an average AUROC of 77.37% ± 5.21 (95% CI: 74.78 to 79.77), which was only slightly superior to the local model's performance of 76.68% ± 3.64 (95% CI: 74.31 to 79.12) [$P = 0.242$]. This limited improvement highlights the pronounced non-IID nature of pediatric chest X-rays compared to adult datasets, even though the Pediatrics dataset had the smallest training size (n = 7,728) among all datasets.

For adult datasets with more representative training data, the conventional FL model often failed to outperform or even match the performance of local training. For the PadChest dataset, FL resulted in a slightly reduced performance with an average AUROC of 84.94% ± 1.16 (95% CI: 84.28 to 85.61) compared to 85.28% ± 0.90 (95% CI: 84.54 to 85.93) for local training [$P = 0.063$]. Similarly, for the CheXpert dataset, which has the largest training size and is more representative, the FL model exhibited significantly inferior performance, achieving an average AUROC of 79.32% ± 7.71 (95% CI: 78.21 to 80.30) compared to 80.99% ± 6.01 (95% CI: 80.09 to 81.89) for local training [$P < 0.001$], (see **Table 2** and **Figure 3**).

These results suggest that the degree of non-IID variability strongly determines the effectiveness of FL. For datasets with high non-IID variability, such as pediatric datasets when combined with adult data, FL struggles to deliver substantial improvements. For large and representative datasets where the training data already captures a broad range of variability, conventional FL in non-IID settings may even degrade performance due to disparities in local data distributions.

**Table 2: Comparison of the diagnostic performance of conventional federated learning (FL) with local training.** The table presents the area under the receiver operating characteristic curve (AUROC) values (expressed as percentages) for the classification of pneumonia and radiographs with no abnormality ("no finding") across two models: local training ("Local") and conventional federated learning ("FL"). Results are reported as mean ± standard deviation (SD) with 95% confidence intervals (CIs). See **Table 1** for further details on dataset characteristics. Differences between Local and FL models were assessed for statistical significance using bootstrapping, and p-values were indicated. Significant differences are indicated in **bold**.

| Test dataset | Training method | No finding [mean ± SD (95% CI)] | Pneumonia [mean ± SD (95% CI)] | Average [mean ± SD (95% CI)] |
|---|---|---|---|---|
| Pediatrics | Local | 73.39 ± 1.39 (70.65, 76.19) | 79.98 ± 1.68 (76.83, 83.05) | 76.68 ± 3.64 (74.31, 79.12) |
| | FL | 72.39 ± 1.41 (69.49, 75.10) | 82.35 ± 1.69 (78.85, 85.51) | 77.37 ± 5.21 (74.78, 79.77) |
| | P-value | 0.173 | **0.031** | 0.242 |
| VinDr-CXR | Local | 90.75 ± 0.59 (89.55, 91.85) | 91.56 ± 0.96 (89.72, 93.34) | 91.15 ± 0.90 (89.92, 92.32) |
| | FL | 93.64 ± 0.45 (92.71, 94.48) | 93.70 ± 0.75 (92.21, 95.18) | 93.67 ± 0.62 (92.74, 94.56) |
| | P-value | **< 0.001** | **0.002** | **< 0.001** |
| ChestX-ray14 | Local | 70.11 ± 0.33 (69.50, 70.78) | 68.28 ± 1.14 (66.11, 70.59) | 69.20 ± 1.24 (68.00, 70.45) |
| | FL | 71.34 ± 0.33 (70.69, 72.00) | 71.70 ± 1.04 (69.67, 73.82) | 71.52 ± 0.79 (70.42, 72.62) |
| | P-value | **< 0.001** | **< 0.001** | **< 0.001** |
| PadChest | Local | 86.04 ± 0.25 (85.56, 86.54) | 84.53 ± 0.65 (83.13, 85.74) | 85.28 ± 0.90 (84.54, 85.93) |
| | FL | 85.99 ± 0.25 (85.49, 86.47) | 83.88 ± 0.64 (82.65, 85.10) | 84.94 ± 1.16 (84.28, 85.61) |
| | P-value | 0.381 | 0.071 | 0.063 |
| CheXpert | Local | 86.96 ± 0.31 (86.36, 87.54) | 75.01 ± 0.85 (73.36, 76.69) | 80.99 ± 6.01 (80.09, 81.89) |
| | FL | 86.99 ± 0.30 (86.37, 87.56) | 71.64 ± 1.00 (69.57, 73.45) | 79.32 ± 7.71 (78.21, 80.30) |
| | P-value | 0.532 | **< 0.001** | **< 0.001** |

## (a) VinDr-CXR dataset

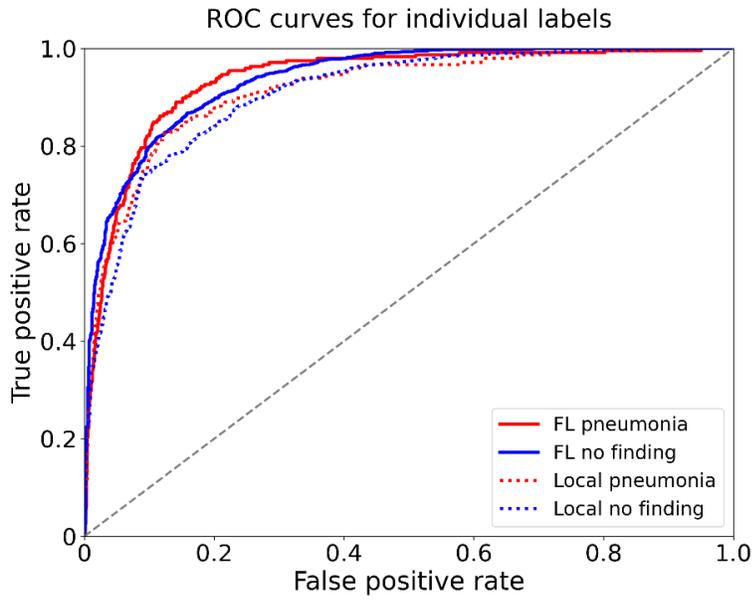
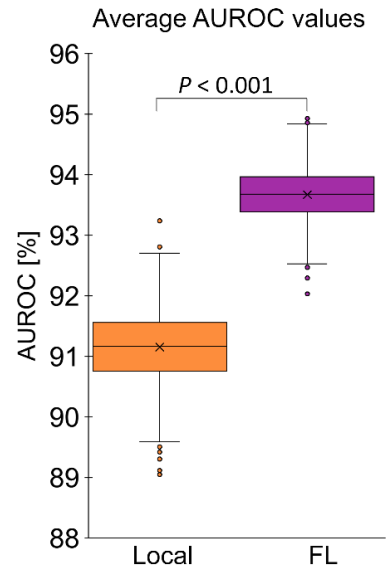

## (b) ChestX-ray14 dataset

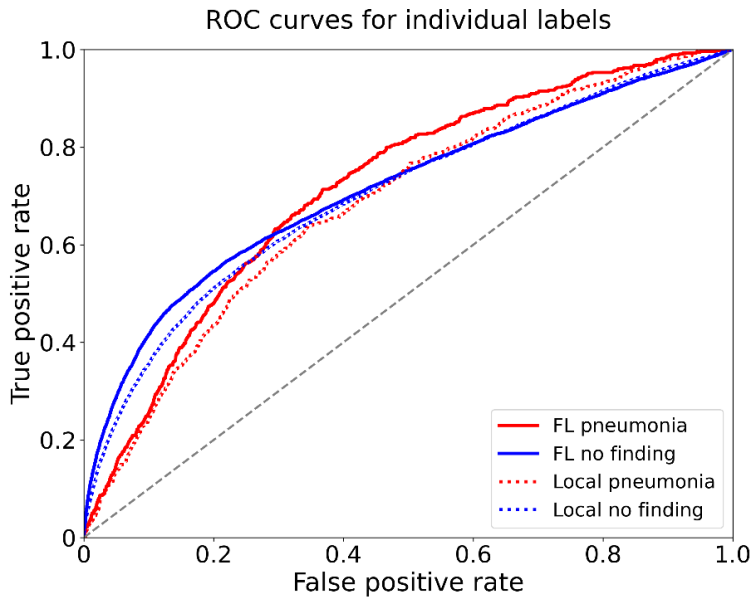
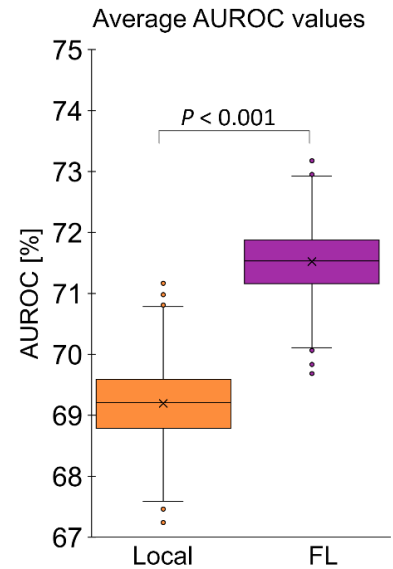

**Figure 2: Diagnostic performance of conventional federated learning (FL) compared to local training for underrepresented adult datasets.** The results present the receiver operating characteristic (ROC) curves for the classification of pneumonia and radiographs with no abnormality ("no finding") along with the average area under the receiver operating characteristic curve (AUROC) values (expressed as percentages) for local training ("Local") and conventional FL ("FL"). Results are shown for **(a)** the VinDr-CXR dataset with n=15,000 training images and n=3,000 test images, and **(b)** the ChestX-ray14 dataset with n=86,524 training images and n=25,596 test images. Statistical significance of the differences between the Local and FL models was evaluated using bootstrapping, with p-values indicated.

### (a) Pediatrics dataset

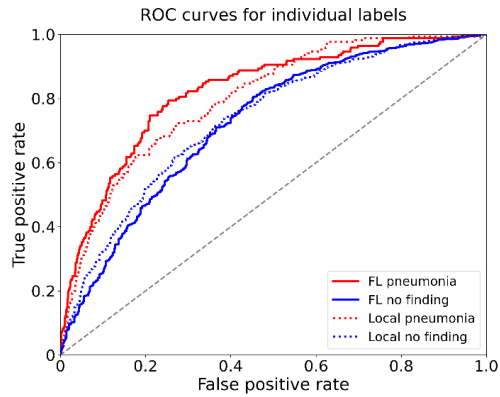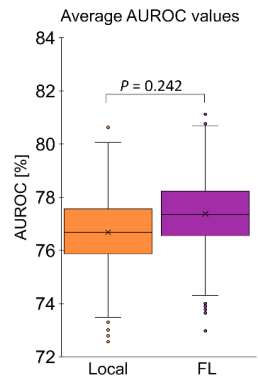

### (b) PadChest dataset

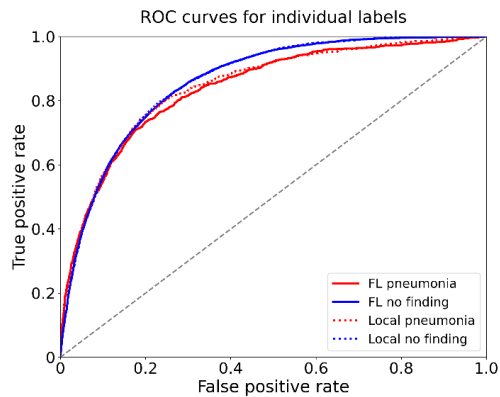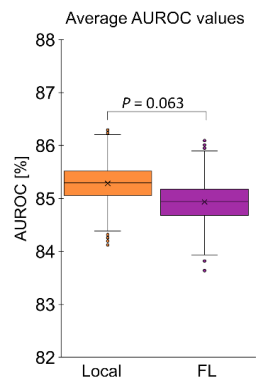

### (c) CheXpert dataset

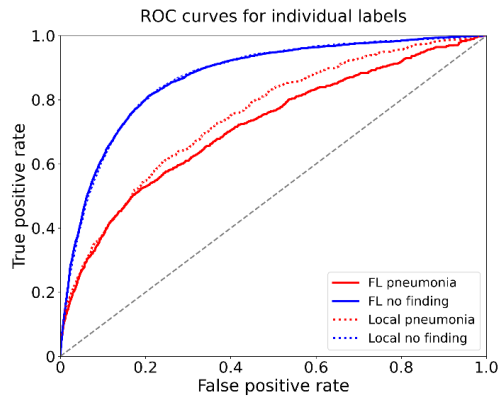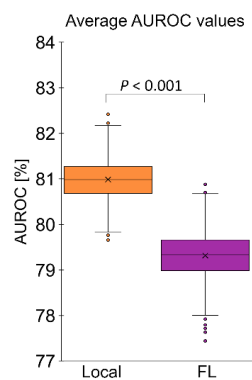

**Figure 3: Diagnostic performance of conventional federated learning (FL) compared to local training for pediatrics and representative adult datasets.** The results present the receiver operating characteristic (ROC) curves for the classification of pneumonia and radiographs with no abnormality ("no finding") along with the average area under the receiver operating characteristic curve (AUROC) values (expressed as percentages) for local training ("Local") and conventional FL ("FL"). Results are shown for **(a)** the Pediatrics dataset with n=7,728 training images and n=1,397 test images, **(b)** the PadChest dataset with n=88,480 training images and n=22,045 test images, and **(c)** the CheXpert dataset with n=128,356 training images and n=29,320 test images. Statistical significance of the differences between the Local and FL models was evaluated using bootstrapping, with p-values indicated.

## 2.3. Self-supervised image representations enable pediatric chest X-rays to benefit from federated learning with adult datasets

Transferring general-purpose image representations from self-supervised learning (SSL) on large-scale non-medical images has been shown to improve performance in chest X-ray analysis[47–49]. Building on this approach, we equipped each local training institution in the FL process with SSL weights derived from the DINOv2 method[46]. For the Pediatrics dataset, this enhancement resulted in an average AUROC of 78.64% ± 6.52 (95% CI: 76.17 to 80.86), which was significantly superior to local training alone (76.68% ± 3.64, 95% CI: 74.31 to 79.12) [$P$ = 0.031], (see **Figure 4**).

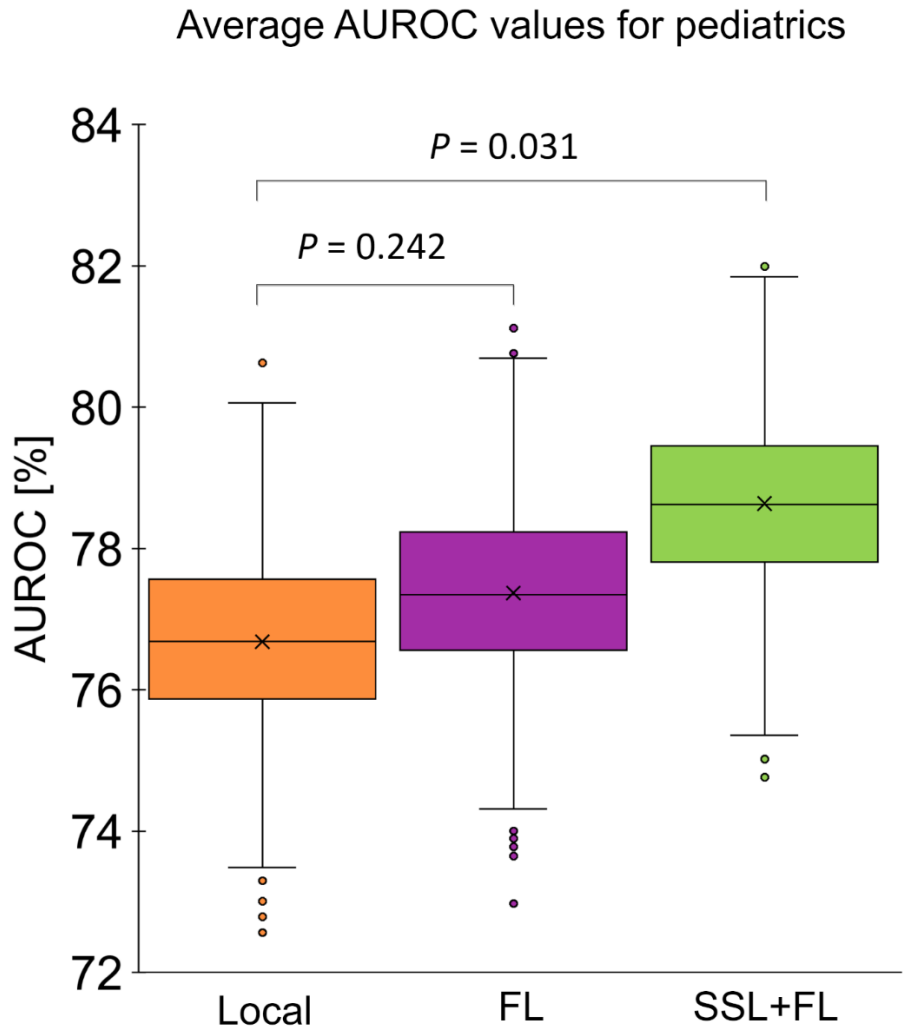

**Figure 4: Diagnostic performance comparison of self-supervised learning (SSL)-based federated learning (FL) with local training for the Pediatrics dataset.** The figure presents the average area under the receiver operating characteristic curve (AUROC) values (expressed as percentages) for the classification of pneumonia and radiographs with no abnormality ("no finding") across three models: local training ("Local"), conventional FL ("FL"), and SSL-based federated learning ("SSL+FL"). Results are shown for the Pediatrics dataset with n=7,728 training images and n=1,397 test images. Statistical significance of the differences between the Local and SSL+FL models was evaluated using bootstrapping, with p-values indicated.

This improvement demonstrates that equipping pediatric data with general-purpose SSL image representations allows them to derive meaningful benefits from FL, even in highly non-IID settings where pediatric and adult datasets are combined.

## 2.4. Self-supervised image representations enhance FL performance in non-IID settings

As shown in **Figure 5**, SSL-based FL significantly outperformed local models in most cases. For datasets with underrepresented training data, such as VinDr-CXR and ChestX-ray14, the SSL-based FL framework maintained the trend observed with conventional FL, delivering significantly higher average AUROC scores compared to local models [$P$ < 0.001 for both] (see **Table 3**).

Importantly, unlike conventional FL, SSL-based FL did not result in any significant diagnostic performance degradations for datasets with representative training data. For the PadChest dataset, where conventional FL slightly reduced performance compared to local training, the SSL-based FL model achieved an average AUROC of 85.82% ± 0.84 (95% CI: 85.13 to 86.52), significantly outperforming the local model's performance of 85.28% ± 0.90 (95% CI: 84.54 to 85.93) [$P$ = 0.007]. Similarly, for the CheXpert dataset, where conventional FL significantly reduced performance compared to local training, the SSL-based FL model achieved an average AUROC of 80.30% ± 7.00 (95% CI: 79.35 to 81.24). This was only slightly inferior to the local model's performance of 80.99% ± 6.01 (95% CI: 80.09 to 81.89) [$P$ = 0.052], further highlighting the capability of SSL-based FL to address non-IID challenges more effectively than conventional FL.

These findings demonstrate that incorporating self-supervised image representations into the FL process mitigate non-IID effects to a great extent, improving performance for both underrepresented and representative datasets.

# 3. Discussion

In this study, we investigated the impact of general-purpose image representations derived from self-supervised learning (SSL) on highly non-independent and identically distributed (non-IID) and heterogeneous federated learning (FL) for large-scale chest X-ray classification. By leveraging SSL representations from the DINOv2[46] framework, we addressed many of the limitations inherent in conventional FL[7]. Using over 400,000 chest X-ray images from both adult and pediatric populations, collected across five diverse datasets from institutions worldwide, we demonstrated that SSL-based FL enhances diagnostic performance in the presence of highly heterogeneous and non-IID data, including scenarios involving collaborative learning between adult and pediatric datasets.

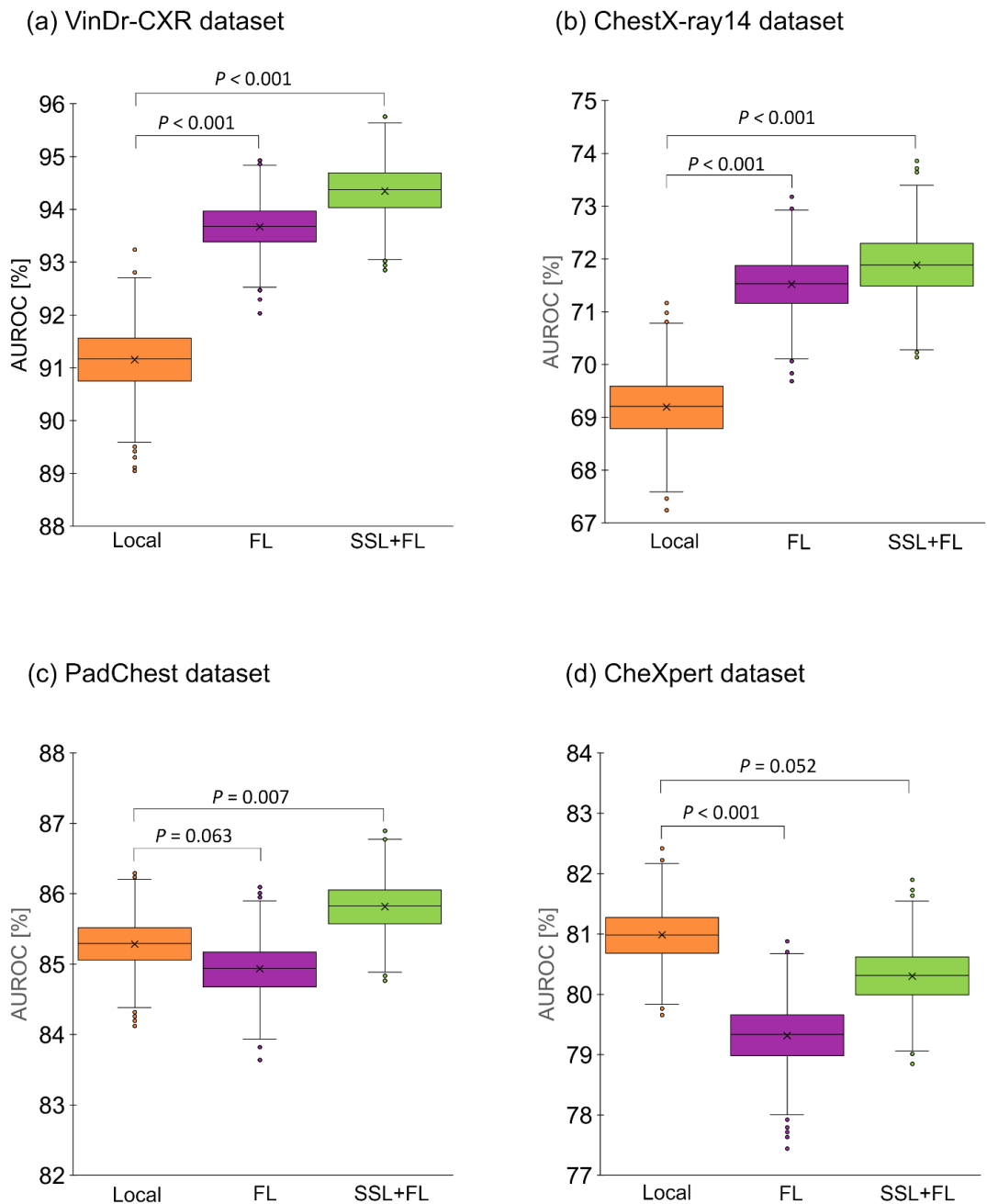

**Figure 5: Diagnostic performance comparison of self-supervised learning (SSL)-based federated learning (FL) with local training for the adult datasets.** The figure presents the average area under the receiver operating characteristic curve (AUROC) values (expressed as percentages) for the classification of pneumonia and radiographs with no abnormality ("no finding") across three models: local training ("Local"), conventional FL ("FL"), and SSL-based federated learning ("SSL+FL"). Results are shown for for **(a)** the VinDr-CXR dataset with n=15,000 training images and n=3,000 test images, **(b)** the ChestX-ray14 dataset with n=86,524 training images and n=25,596 test images, **(c)** the PadChest dataset with n=88,480 training images and n=22,045 test images, and **(d)** the CheXpert dataset with n=128,356 training images and n=29,320 test images. Statistical significance of the differences between the Local and SSL+FL models was evaluated using bootstrapping, with p-values indicated.

**Table 3: Comparison of the diagnostic performance of self-supervised learning (SSL)-based federated learning (FL) with local training.** The table presents the area under the receiver operating characteristic curve (AUROC) values (expressed as percentages) for the classification of pneumonia and radiographs with no abnormality ("no finding") across two models: local training ("Local") and SSL-based federated learning ("SSL+FL"). Results are reported as mean ± standard deviation (SD) with 95% confidence intervals (CIs). See **Table 1** for further details on dataset characteristics. Differences between Local and SSL+FL models were assessed for statistical significance using bootstrapping, and p-values were indicated. Significant differences are indicated in **bold**.

| Test dataset | Training method | No finding [mean ± SD (95% CI)] | Pneumonia [mean ± SD (95% CI)] | Average [mean ± SD (95% CI)] |
|---|---|---|---|---|
| Pediatrics | Local | 73.39 ± 1.39 (70.65, 76.19) | 79.98 ± 1.68 (76.83, 83.05) | 76.68 ± 3.64 (74.31, 79.12) |
|  | SSL+FL | 72.29 ± 1.46 (69.30, 75.03) | 84.98 ± 1.57 (81.72, 87.78) | 78.64 ± 6.52 (76.17, 80.86) |
|  | P-value | 0.185 | **< 0.001** | **0.031** |
| VinDr-CXR | Local | 90.75 ± 0.59 (89.55, 91.85) | 91.56 ± 0.96 (89.72, 93.34) | 91.15 ± 0.90 (89.92, 92.32) |
|  | SSL+FL | 94.63 ± 0.43 (93.68, 95.43) | 94.07 ± 0.78 (92.48, 95.53) | 94.35 ± 0.69 (93.30, 95.26) |
|  | P-value | **< 0.001** | **< 0.001** | **< 0.001** |
| ChestX-ray14 | Local | 70.11 ± 0.33 (69.50, 70.78) | 68.28 ± 1.14 (66.11, 70.59) | 69.20 ± 1.24 (68.00, 70.45) |
|  | SSL+FL | 71.76 ± 0.34 (71.09, 72.44) | 72.00 ± 1.11 (69.76, 74.22) | 71.88 ± 0.83 (70.72, 73.08) |
|  | P-value | **< 0.001** | **< 0.001** | **< 0.001** |
| PadChest | Local | 86.04 ± 0.25 (85.56, 86.54) | 84.53 ± 0.65 (83.13, 85.74) | 85.28 ± 0.90 (84.54, 85.93) |
|  | SSL+FL | 86.50 ± 0.25 (86.01, 86.97) | 85.14 ± 0.66 (83.80, 86.40) | 85.82 ± 0.84 (85.13, 86.52) |
|  | P-value | **< 0.001** | 0.927 | **0.007** |
| CheXpert | Local | 86.99 ± 0.30 (86.37, 87.56) | 71.64 ± 1.00 (69.57, 73.45) | 79.32 ± 7.71 (78.21, 80.30) |
|  | SSL+FL | 87.26 ± 0.31 (86.68, 87.88) | 73.34 ± 0.92 (71.50, 75.10) | 80.30 ± 7.00 (79.35, 81.24) |
|  | P-value | **0.019** | **0.027** | 0.052 |

Our findings underscore the critical influence of non-IID variability on FL performance. Conventional FL struggled to generalize effectively in most of the non-IID settings, particularly for larger datasets with more representative training distributions, such as CheXpert[33] and PadChest[34]. For the pediatric dataset[25], where variability is more pronounced, FL offered only marginal improvements over local training, underscoring the limitations of conventional FL in handling highly heterogeneous datasets. These results align with prior research[22] indicating that non-IID variability exacerbates performance imbalances, particularly when datasets differ in size, labeling, or demographic characteristics[17,18,41,50].

The integration of SSL weights into FL transformed its performance by introducing robust, general-purpose feature representations pre-trained on diverse, large-scale datasets[47]. This enhancement was particularly pronounced for pediatric data, where SSL-based FL achieved significant improvements compared to local training. These results highlight the ability of SSL-based approaches to bridge performance gaps for underrepresented populations, offering a practical solution to inequities often observed in real-world medical imaging scenarios. For adult datasets, SSL-based FL mitigated the performance degradation observed with conventional FL in large datasets such as CheXpert, where diagnostic accuracy improved from significantly lower levels to results that were no longer statistically different from local training, albeit still slightly lower. Additionally, SSL-based FL maintained or improved performance in smaller datasets like VinDr-CXR[32]. These findings demonstrate that SSL weights harmonize disparate data distributions, enabling better collaboration across institutions with varied datasets and addressing the broader challenges of non-IID variability.

Beyond mitigating non-IID variability, SSL-based FL illustrates its potential to advance diagnostic AI in healthcare. By preserving data privacy through FL while leveraging the scalability and generalization capabilities of SSL, this approach aligns with the ethical and logistical demands of modern AI development. The ability to collaborate globally across institutions, leveraging diverse data sources without compromising patient confidentiality, represents a step toward scalable and equitable AI solutions for medical imaging.

Our study has several limitations. First, while the pediatric dataset used in this analysis is the largest publicly available pediatric chest X-ray dataset to date[25], it remains relatively small compared to the adult datasets. This disparity reflects the real-world challenge of acquiring labeled pediatric data. Future studies should explore the scalability of SSL-based FL frameworks using larger and more diverse pediatric datasets to confirm broader applicability and to better capture the variability within pediatric populations. Second, the collaborative training in this study was simulated within a single institution's network. By isolating computing entities for each virtual site participating in the collaborative training process, we emulated a practical federated learning scenario where updates from multiple sites are aggregated at a central server[22,23]. This setup ensured that comparisons were inherently paired, with identical hyperparameters used across datasets and training strategies. While this controlled environment does not fully replicate the complexities of real-world FL—such as network latency, computational resource heterogeneity, and geographic distribution—it does not affect the diagnostic performance outcomes, as the underlying model updates and evaluation metrics remain unchanged. Future work should implement FL in real-world multi-institutional setups to assess the operational challenges and their potential impact on procedural efficiency. Third, while we performed strictly paired comparisons to systematically assess the effects of SSL in FL, we used 224×224 pixel inputs for model training. Although prior studies suggest that resolutions of 256×256 or higher are sufficient for chest radiograph classification using convolutional networks[51,52], the use of 224×224 aligns with the ViT architecture's capabilities[22,47,53,54], and the availability of DINOv2 weights trained for this input size[46]. Future research should prioritize developing SSL weights for higher-resolution inputs to enable a more comprehensive evaluation of whether increased resolutions provide additional benefits in FL for medical imaging. Fourth, this study focused on two diagnostic labels—pneumonia and no abnormality—due to their consistent availability across datasets. While these labels are clinically significant and widely studied, they represent a limited scope. Future research should expand this approach to include more diagnostic categories and extend beyond radiographs to domains such as gigapixel pathology imaging[55] and three-dimensional volumetric medical imaging (e.g., magnetic resonance imaging[56]). This would provide a more comprehensive evaluation of the broader applicability of SSL-based FL frameworks in medical imaging.

In conclusion, this study demonstrates that general-purpose self-supervised representations can effectively address the limitations of conventional FL in highly non-IID settings, particularly for collaborative learning involving both adult and pediatric data. By significantly enhancing diagnostic performance, SSL-based FL emerges as a robust, scalable, and privacy-preserving framework for advancing AI development in healthcare. These findings lay the groundwork for future research to further refine FL frameworks equipped with general-purpose self-supervised representations from non-medical images, ultimately aiming to improve diagnostic accuracy and patient outcomes across diverse clinical environments.

# 4. Online methods

## 4.1. Ethics statement

This study was conducted in compliance with all applicable local and national guidelines and regulations. The data utilized in this research were sourced from previously published studies and are publicly accessible. As the study did not involve human subjects or patients, it was exempt from institutional review board approval and did not require informed consent.

## 4.2. Patient datasets

This study included a total of n=407,648 frontal chest radiographs collected from various institutions worldwide. Patient ages ranged from 0 (less than 6 months) to 105 years. The median ages for patients in the adult datasets—VinDr-CXR[32], ChestX-ray14[36], PadChest[34], CheXpert[33]—were 57, 49, 62, and 61 years, respectively, while the median age for the pediatric dataset[25] was 2 years. Detailed characteristics of all datasets used in this study are summarized in **Table 1**. Below, we provide a brief description of each dataset included in the analysis.

### 4.2.1. Pediatrics dataset

The Pedi-CXR[25] (also known as VinDr-PCXR[57]) dataset is the largest publicly available pediatric chest X-ray dataset with labeled studies to date. It comprises 9,125 posteroanterior radiographs from children under the age of 10 (median age: 2 years) in Vietnam. All radiographs were manually annotated by a team of three radiologists, each with at least 10 years of experience. For this study, we utilized the dataset's original split, with n=7,728 images in the training set and n=1,397 images in the test set, as provided by the dataset authors.

### 4.2.2. VinDr-CXR dataset

The VinDr-CXR[32] dataset consists of a curated subset of 18,000 images selected from over 100,000 chest radiographs collected at two Vietnamese hospitals. These images were captured using a diverse array of medical imaging devices from multiple manufacturers. The imaging findings were meticulously labeled by a team of 17 expert radiologists, with each image independently annotated by three radiologists. Labeling was based on the frequency and visibility of conditions in chest radiographs. For this study, we utilized the original training set (n=15,000) and test set (n=3,000) as provided by the dataset[32].

### 4.2.3. ChestX-ray14 dataset

The ChestX-ray14[36] dataset, provided by the National Institutes of Health, focuses on identifying 14 common thoracic pathologies with guidance from radiologists in selecting these conditions. The image labeling process was performed automatically using natural language processing[58], which identified the presence or absence of specific pathologies while carefully addressing negations and uncertain terms. Labeling was conducted in two stages[36]. First, disease concepts were extracted primarily from specific sections of the radiology reports. Second, reports showing no evidence of pathology were categorized as "no finding." The dataset contains a total of 112,120 radiographs, without an officially defined training/test split[47]. For this study, we performed a patient-wise random split of approximately 80%/20%, resulting in n=86,524 images for the training set and n=25,596 images for the test set (see **Table 1**).

### 4.2.4. PadChest dataset

The PadChest[34] dataset, collected in Spain, includes 109,931 studies, resulting in 110,525 frontal chest radiographs. Of these, 27% of the reports (27,593 studies) were manually reviewed and labeled by expert radiologists. The manually labeled subset was then used to train a multilabel text classifier, which was subsequently applied to automatically annotate the remaining 73% of the reports [29]. For this study, we performed a patient-wise random split, balanced between the manually labeled and automatically labeled radiographs, with an approximate 80%/20% distribution[34,47]. This resulted in n=88,480 images for the training set and n=22,045 images for the test set (see **Table 1**).

### *4.2.5. CheXpert dataset*

The CheXpert[33] dataset contains 224,316 frontal and lateral chest radiographs from 65,240 patients, collected at Stanford Hospital. Observations were labeled using a rule-based natural language processing system guided by radiologists. Key findings were extracted from the Impression section of radiology reports, categorized as negative, uncertain, or positive. Ambiguous or explicitly uncertain mentions were labeled as "uncertain," while mentions without clear classification were defaulted to positive[47]. Observations not mentioned in the reports were left blank[33]. For this study, a patient-wise random split of approximately 80%/20% was performed, resulting in n=128,356 images for the training set and n=29,320 images for the test set (see **Table 1**).

For all datasets, reports were labeled "no finding" if no disease was detected or if the report explicitly indicated normal findings. There was strictly no patient overlap between the training and test sets.

## 4.3. Data pre-processing

The only common labels across all adult and pediatric datasets were "pneumonia" and "no finding," making these the primary diagnostic labels of interest. Following approaches from previous studies[22,23,37,47,59,59], we employed a binary classification method, categorizing each radiograph as either positive or negative. The pediatric dataset, VinDr-CXR, ChestX-ray14, and PadChest datasets were already binary-labeled. For the CheXpert dataset, "certain negative" and "uncertain" labels were grouped as "negative," while only "certain positive" labels were classified as "positive," in line with prior research.

A unified image pre-processing workflow was employed[22,23,47,60]. Chest x-rays were first resized to a standard resolution of 224×224 pixels to ensure compatibility with the model architecture. Intensity values were normalized using min-max scaling to bring all images into a comparable range, improving model convergence[35]. To further refine the visual quality of the images, histogram equalization[23,35] was applied, enhancing contrast and emphasizing key features. Pre-processing was performed independently at each participating institution, adhering to a standardized protocol to maintain consistency within the federated learning framework. To enrich the dataset and improve model robustness, data augmentation techniques were applied, including random rotations of up to 10 degrees and random horizontal flips.

## 4.4. Experimental design

To ensure benchmarking consistency and facilitate strictly paired comparisons across different experiments, the unified image pre-processing workflow was applied uniformly to all datasets. Additionally, the training and test sets for each of the five datasets remained fixed throughout the study and all experiments (see **Table 1** for dataset statistics). This setup resulted in five held-out test sets with sample sizes of n=1,397 (Pediatrics), n=3,000 (VinDr-CXR), n=25,596 (ChestX-ray14), n=22,045 (PadChest), and n=29,320 (CheXpert).

As the baseline scenario, we modeled a realistic and commonly observed setup where each institution independently trains a diagnostic model locally, using only its own data without collaboration. For this scenario, separate diagnostic models were trained for multi-label classification of "pneumonia" and "no finding" using the training sets from each dataset: n=7,728 (Pediatrics), n=15,000 (VinDr-CXR), n=86,524 (ChestX-ray14), n=88,480 (PadChest), and n=128,356 (CheXpert). Importantly, identical network architectures and training procedures were employed across all local institutions to ensure fairness in comparisons.

Next, a conventional FL scenario was implemented, with each dataset serving as an independent local site (details provided below). Training in this setup was performed using the same training sets as in the baseline scenario.

Finally, in the main experimental scenario, each institution was equipped with self-supervised representations. Instead of standard FL initialization, training at each site began from these pre-trained self-supervised parameters (details provided below).

The resulting networks from all three scenarios—baseline ("Local"), conventional FL ("FL"), and self-supervised representation-equipped FL ("SSL+FL")—were evaluated on the fixed held-out test sets to ensure robust and consistent performance comparisons.

## 4.5. Network architecture

The network architecture used in this study was based on the original 12-layer vision transformer (ViT) implementation proposed by Dosovitskiy et al.[45] The model processes input images of dimensions 224×224×3, organized into batches of size 16. The initial embedding layer uses a 16×16 or 14x14 convolutions with a strides of 16×16 and 14x14, effectively dividing each image into a sequence of non-overlapping patches. Each patch is then flattened and mapped to a 768-dimensional vector using a learnable linear transformation. These embeddings are supplemented with positional encodings, resulting in a sequence of $T$ vectors. This sequence is then passed to the transformer[61] encoder.

The encoder comprises 12 layers, each including a multi-head self-attention mechanism[61] and a feed-forward network. The self-attention mechanism computes the attention scores based on the query ($Q$), key ($K$), and value ($V$) matrices derived from the input embeddings:

$$Attention(Q,K,V) = softmax\left(\frac{QK^T}{\sqrt{d_k}}\right)V, \qquad (1)$$

where $d_k$ is the dimensionality of the key vectors, and the softmax function ensures that attention scores sum to 1. The outputs from the self-attention module are passed through the feed-forward network, which includes two fully connected layers with a ReLU activation[62]. The hidden layer size of the feed-forward network is 3,072.

The output of the transformer encoder is passed to a multi-layer perceptron classification head. Since this study focuses on multilabel binary classification tasks (i.e., classifying each image for the presence or absence of "pneumonia" and "no finding"), the classification head outputs one logit per label. A sigmoid activation function is applied to each logit to convert them into probabilities.

The network consists of approximately 86 million trainable parameters and was initialized with ImageNet-21K[63] pretrained weights. The loss function used for training was a binary weighted cross-entropy, where the weight for each class was inversely proportional to its frequency in the training data. This approach addresses the imbalance in class distribution by assigning higher importance to underrepresented labels during training. Optimization was performed using the AdamW[64] optimizer with a learning rate of $1 \times 10^{-5}$. Hyperparameters were systematically tuned to achieve consistent convergence and robust performance across all experiments.

## 4.6. Federated learning

The federated learning (FL) framework employed in this study was based on the Federated Averaging (FedAvg) algorithm, a widely used approach introduced by McMahan et al.[7] This method facilitates collaborative training across multiple institutions while preserving data privacy and security. In this setup, five participating institutions—Pediatrics, VinDr-CXR, ChestX-ray14, PadChest, and CheXpert—trained local models independently using their respective datasets. Each local model, denoted as $w_i^t$, was trained on the institution's dataset $D_i$ during the $t$-th round of training. A single training round was defined as one epoch of training on the entire local dataset at each site. The objective of local training at each site was to minimize a loss function $L_i(w)$, defined as the average loss over the institution's dataset, according to the following equation,

$$L_i(w) = \frac{1}{|D_i|} \sum_{(x,y) \in D_i} \ell(f_w(x), y), \qquad (2)$$

where $\ell$ is the loss function (cross-entropy loss in this study), $f_w(x)$ is the model's prediction for input $x$, and $y$ is the ground truth label. Parameters at each local site were updated using the AdamW[64] optimizer, which incorporates weight decay for better generalization. The parameter update rule is given by the following equation,

$$w_i^{t+1} = w_i^t - \eta \cdot \text{AdamW}(\nabla L_i(w_i^t)), \qquad (3)$$

where $\eta$ denotes the learning rate. After completing local training, each institution transmitted its updated parameters $w_i^{t+1}$ to a central server. The server aggregated these parameters to produce a global model $w^{t+1}$ as follows,

$$w^{t+1} = \frac{1}{N} \sum_{i=1}^{N} w_i^{t+1}, \qquad (4)$$

where $N$ is the total number of participating institutions ($N = 5$ in this study). The training dataset sizes were n=7,728 for Pediatrics, n=15,000 for VinDr-CXR, n=86,524 for ChestX-ray14, n=88,480 for PadChest, and n=128,356 for CheXpert.

The updated global model was redistributed to all participating institutions, where it served as the initialization for the next round of training. This iterative process continued until the global model converged. After convergence, the final global model was distributed to each institution for evaluation. Each institution used this model to independently assess performance on its respective test dataset.

## 4.7. General-purpose self-supervised representations for federated learning

The self-supervised learning (SSL) approach in this study utilized general-purpose image representations generated by the DINOv2[46] framework, developed by Meta AI. DINOv2 represents an advancement of the DINO[65] method, focusing on extracting diverse and robust visual features from large-scale datasets. The underlying dataset for DINOv2 consisted of 142 million unique images curated from a variety of sources such as Google Landmarks[66] and other public and internal web repositories, ensuring a broad and diverse representation of visual concepts[47].

The DINOv2 training framework employs ViT[45] architectures. Self-supervised training of DINOv2 synthesizes elements from various state-of-the-art SSL methodologies, including DINO[65], iBOT[67], and SwAV[68]. It incorporates two primary loss objectives: the image-level objective and the patch-level objective.

The image-level objective ensures consistency between representations of different augmented views of the same image. For an image $x$, two augmented views $x_s$ (student) and $x_t$ (teacher), are processed through a ViT. The teacher network's parameters are updated using an exponential moving average of the student network's parameters. The image-level loss is computed as the following equation,

$$L_{image} = -\sum_{k} p_t\left(x_t^{(k)}\right) \log p_s\left(x_s^{(k)}\right), \quad (5)$$

where $p_t\left(x_t^{(k)}\right)$ and $p_s\left(x_s^{(k)}\right)$ are the output probabilities of the teacher and student networks, respectively, for feature dimension $k$.

The patch-level objective focuses on learning localized features by leveraging selective masking. Certain input patches are masked for the student network, and the features of the remaining patches are compared to the corresponding features from the teacher network. The patch-level loss is computed as the following equation,

$$L_{image} = -\sum_{j}\sum_{k} p_t\left(x_t^{(j,k)}\right) \log p_s\left(x_s^{(j,k)}\right), \quad (6)$$

where $j$ indexes the patches, and $x_t^{(j,k)}$ and $x_s^{(j,k)}$ represent the teacher and student features for patch $j$ and dimension $k$.

The total DINOv2 loss is a weighted combination of the image-level and patch-level objectives,

$$L_{DINOv2} = \alpha L_{image} + \beta L_{patch}, \tag{7}$$

where $\alpha$ and $\beta$ are weights balancing the contributions of the two objectives[46].

### 4.7.1. SSL training

To enhance feature distribution, Sinkhorn-Knopp[69] normalization and KoLeo[70] regularization were applied[71]. Training was conducted at a resolution of 224×224 pixels for most iterations, with the resolution increased to 416×416 pixels in the final iterations. This hybrid resolution strategy optimized computational efficiency while maintaining high performance[68,72]. For further details on the DINOv2 methodology, refer to the original publication[46].

## 4.8. Evaluation

The primary evaluation metric was the area under the receiver operating characteristic curve (AUROC), supplemented by additional evaluation metrics such as accuracy, specificity, and sensitivity. The thresholds were chosen according to the Youden's criterion[73].

### 4.8.1. Statistical analysis

We analyzed the AI models using Python v3.9 and SciPy v1.10, NumPy v1.23, and scikit-learn v1.2 libraries. The quantitative evaluation metrics are represented as mean ± standard deviation (with 95% confidence interval values stated). We employed bootstrapping[74] with replacement and 1,000 redraws in the test sets to determine the statistical spread and whether AUROC values differed significantly. A p-value < 0.05 was considered significant.

## 4.9. Code availability

All source codes for federated learning, self-supervised transfer learning, training and evaluation of the networks, statistical analysis, data augmentation, image analysis, and pre-processing are publicly available at https://github.com/mahshadlotfinia/FLTLCXR. All code for the experiments was developed in Python v3.9 using the PyTorch v2.0 framework.

## 4.10. Data availability

The datasets used in this study are available as follows: The ChestX-ray14 and PadChest datasets are publicly accessible via https://www.v7labs.com/open-datasets/chestx-ray14 and https://bimcv.cipf.es/bimcv-projects/padchest/, respectively. The VinDr-CXR and the pediatrics datasets are restricted-access resources and can be accessed through PhysioNet upon agreement to the relevant data protection requirements at https://physionet.org/content/vindr-cxr/1.0.0/ and https://physionet.org/content/vindr-pcxr/1.0.0/, respectively. Access to the CheXpert dataset can be requested at https://stanfordmlgroup.github.io/competitions/chexpert/.

## 4.11. Author contributions

ML and STA designed the study and performed the formal analysis. The manuscript was written by ML, AT, and STA, and reviewed and corrected by STA. The experiments were performed by ML and AT. The software was developed by ML and STA. The statistical analyses were performed by ML. STA provided clinical expertise. ML, AT, SS, MJ, and STA provided technical expertise, read the manuscript, and agreed to the submission of this paper.